\documentclass[11pt,a4paper]{article}
\usepackage[hyperref]{emnlp2020}
\usepackage{times}
\usepackage{latexsym}
\usepackage{graphicx}

\usepackage[linesnumbered,ruled,vlined]{algorithm2e}
\usepackage{microtype}
\usepackage{amsfonts,amssymb}
\usepackage{amsmath}
\usepackage{booktabs}
\usepackage{multirow}
\usepackage{bigstrut,bigdelim}
\usepackage{paralist}
\usepackage{bbm}
\usepackage{color}
\usepackage{makecell}
\usepackage{url}
\usepackage{subcaption, caption}
\usepackage{hyperref}
\usepackage[hyphenbreaks]{breakurl}

\aclfinalcopy 

\title{Are Pre-trained Language Models Knowledgeable to Ground Open Domain Dialogues? }

\author{
Yufan Zhao$^1$, Wei Wu$^2$, Can Xu$^1$,\\
$^1$Microsoft Corporation, Beijing, China\\
$^2$Meituan, Beijing, China \\
\texttt{\{yufzhao,caxu\}@microsoft.com} \\
\texttt{wuwei19850318@gmail.com}\\
}

\date{}

\begin{document}
\maketitle

\begin{abstract}
We study knowledge-grounded dialogue generation with pre-trained language models. Instead of pursuing new state-of-the-art on benchmarks, we try to understand if the knowledge stored in parameters of the pre-trained models is already enough to ground open domain dialogues, and thus allows us to get rid of the dependency on external knowledge sources in generation. Through extensive experiments on benchmarks, we find that by fine-tuning with a few dialogues containing knowledge, the pre-trained language models can outperform the state-of-the-art model that requires external knowledge in automatic evaluation and human judgment, suggesting a positive answer to the question we raised.

\end{abstract}
       \section{Introduction}
While techniques of open domain dialogue generation \cite{vinyals2015neural,xing2017hierarchical,zhang2019recosa} have been applied in industrial products \cite{shum2018eliza,ram2018conversational}, 
people can still feel the gap between the dialogue systems and humans, especially when they dive into a specific topic of interest. To bridge the gap, researchers consider grounding open domain dialogues by external knowledge which could be retrieved either from structured knowledge bases \cite{zhou2018commonsense,moon2019opendialkg,tuan2019dykgchat} or from unstructured documents \cite{dinan2018wizard}. 
Dialogue generation now is based on both conversation contexts and the external knowledge which hints the generation model how to go deep for the topic in discussion. 

In this work, we investigate if a large scale pre-trained language model can instead serve as a knowledge base in open domain dialogue generation. The work is motivated by two lines of research emerging recently: (1) interestingly, some recent studies indicate that pre-trained language models have packed enough knowledge in their parameters, and thus they can do a good job in question-answering tasks without the need of access to external knowledge \cite{petroni2019language,roberts2020much}. Thus, we are curious if similar results can be achieved in open domain dialogue generation. If the answer is ``yes'', then we can get rid of the dependency on external knowledge sources, and obtain a simpler and more flexible architecture with a better generalization ability inherited from pre-training with massive text corpus \cite{radford2019language};  and (2) pre-training techniques have exhibited compelling performance on the task of open domain dialogue generation \cite{zhang2019dialogpt,wolf2019transfertransfo}. Particularly, a recent paper \cite{zhang2019dialogpt} has demonstrated with examples that a pre-trained generation model can reply with commonsense knowledge. Therefore, it is interesting to further explore to what extend a pre-trained language model can keep conversation smooth, knowledgeable, and reasonable with in-depth analysis.

The test beds are benchmarks of knowledge-grounded dialogue generation, including Wizard of Wikipedia (Wizard) \cite{dinan2018wizard}, CMU Document Grounded Conversations (CMU$\_$DoG) \cite{zhou2018dataset}, and Topical-Chat (TC) \cite{gopalakrishnan2019topical}, in which we discard the external knowledge passages. The remaining dialogues are basically deep discussions about specific topics between two participants, and thus are suitable probes. 
We choose DialoGPT \cite{zhang2019dialogpt}, GPT-2$_{finetune}$, and DialoGPT$_{finetune}$ as the pre-trained language models for investigation, where  GPT-2$_{finetune}$ and DialoGPT$_{finetune}$ refer to the OpenAI GPT-2 model \cite{radford2019language} and DialoGPT fine-tuned on the training data of the benchmarks respectively. Evaluation results on both automatic metrics and human judgment indicate that without the aid of external knowledge, the fine-tuned models are still capable of replying with proper and specific content based on the knowledge encoded in its parameters, though sometimes they may make mistakes on details.

Our contributions include: (1) the first systematic study on the possibility of using a pre-trained language model as a knowledge base to ground open domain dialogues; and (2) insights and ideas for future work from extensive experiments.  

	   \section{Pre-trained Language Models}
We choose GPT-2 as the backbone of the pre-trained language models. 
Though the models have exhibited strong performance on a variety of language generation tasks \cite{radford2019language}, they are not suitable for dialogue generation under a zero-shot setting, as the original models often synthesize a long paragraph after a conversation context. For example, on the test sets of Wizard, we observe that ``responses'' generated by the $24$-layer GPT-2 are made up of 402 tokens on average, which cannot be regarded as conversational replies any more\footnote{We have tried several heuristics, such as setting up an upper bound for the length of the responses, as remedies, but in general the heuristics will make the responses ungrammatical.}.  Therefore, we alternatively consider the following variants of GPT-2 which are well adapted to dialogues.

\paragraph{DialoGPT.}  The model follows the architecture of OpenAI GPT-2, and is trained (either from scratch or from OpenAI GPT-2) with $147$M Reddit dialogues \cite{zhang2019dialogpt}. We choose the model trained from OpenAI GPT-2 with $345$M parameters, as it shows the best performance in the evaluation in \cite{zhang2019dialogpt}. The model is implemented based on the code shared at \url{https://github.com/microsoft/DialoGPT}.   

\paragraph{GPT-2$_{finetune}$.} We fine-tune the OpenAI GPT-2 model with $345$M parameters on the training data of the benchmarks. The model removes the influence of the Reddit data, and the fine-tuning step biases the distribution of language to the conversations of the crowd-workers who participate in the construction of the benchmarks. 

\paragraph{DialoGPT$_{finetune}$.} We further fine-tune DialoGPT ($345$M) with the training data of the benchmarks.  Through a comparison with DialoGPT, we can look into how the small training data affects the capability of the pre-trained model in terms of responding with knowledge. 
	   \section{Experiments}
\subsection{Datasets}
All the three benchmark datasets are built with crowd-sourcing on Amazon Mechanical Turk, and are split into training sets, validation sets, and test sets by the data owners. In Wizard and CMU$\_$DoG, knowledge is retrieved from Wikipedia, while in TC, besides wiki articles, Washington Post articles and Reddit fun facts are also utilized as the knowledge sources. Unlike CMU$\_$DoG that focuses on the movie domain, both Wizard and TC cover a wide range of topics from multiple domains. 
Various configurations are set up to simulate conversation scenarios in real world. In Wizard, a wizard tells an apprentice about what he/she learns from the knowledge about a specific topic. In addition to wizard-apprentice conversations, CMU$\_$DoG also contains conversations between two workers who know the background documents and try to discuss the content in depth. In TC, participants play symmetric and asymmetric roles according to the knowledge they can access under $5$ settings. Dialogues under different configurations are combined in each dataset, and only the turns where knowledge is accessible are considered in response generation. 
Wizard and TC further split the test sets into Seen/Frequent and Unseen/Rare where the former contains topics frequently appearing in the training sets and the latter contains topics infrequently or never appearing in the training sets. More details can be found in the supplementary material. 

\subsection{Implementation Details}
Both GPT-2$_{finetune}$ and DialoGPT$_{finetune}$ are trained on 4 Nvidia V100 machines until there is no progress in validation loss. We adopt Adam algorithm \cite{kingma2014adam} in optimization with a learning rate 1e-5 as in \cite{zhang2019dialogpt}. For all the three models, we use the vocabulary in OpenAI GPT-2 and DialoGPT that consists of $50,257$ entries. The latest $128$ tokens in a context are kept as input. 

\begin{table*}[!t]
		\small
		\centering	
		{		
			\begin{tabular}{c|l}
				\hline
				\multirow{1}{*}{}   &\textbf{Case-1}\\ \hline
				\multirow{2}{*}{Context}   &\textbf{Utter1}: My favorite color is blue! It is important when considering things like decoration and art. 
				\\ &\textbf{Utter2}: Blue is a great color especially in the summer\\ \hline
				\multirow{1}{*}{DialoGPT} &It is !\\ \hline
				\midrule
				\multirow{1}{*}{}   &\textbf{Case-2}\\ \hline
				\multirow{2}{*}{Context}   &\textbf{Utter1}: Do you like Elvis Presley? he was a famous american singer
				\\ &\textbf{Utter2}: He is pretty good, I am from Nashville, So i've even been to graceland.
				\\  \hline
				\multirow{1}{*}{DialoGPT$_{finetune}$.} & He died in 1969, where did you live when you visited? \textcolor{red}{(wrong time)}\\ \hline
				\multirow{1}{*}{GPT$_{finetune}$.} & He moved to Nashville, Tennessee when he was 14 in 1926. \textcolor{red}{(wrong time)}
                \\\hline
				\midrule
				\multirow{1}{*}{}   &\textbf{Case-3}\\ \hline
				\multirow{3}{*}{Context}   &\textbf{Utter1}: I am an accountant. Thanks to the establishment of the Italian mathematician Luca Pacioli \\ &\qquad \ \ \ \ \,  in 1494.
				\\ &\textbf{Utter2}: Oh? How was Luca Pacioli instrumental in the establishment of accountancy?
				\\ \hline
				\multirow{2}{*}{GPT$_{finetune}$.} & He developed the concept of accounting in 1795. I am the customer of the banks and I provide the \\ &  service of being in the control of account receivables and inventory.\\ \hline
				
			\end{tabular}
		}
		\caption{\label{case} Case study.}
	\end{table*}

\begin{table*}[!t]
    \small
    \centering
      \scriptsize
       \scalebox{0.9}{
        \begin{tabular}{clcccccccccccc}
            \toprule
            Dataset  & Model  & PPL & F1 & BLEU-1 & BLEU-2 & BLEU-3 & BLEU-4 & Average & Extrema  & Greedy  \\
            \midrule
            \multirow{5}{*}{Wizard-of-Wikipedia(Test Seen)} 
            &TMN & 66.5 & 15.9 & 0.184 & 0.073 & 0.033 & 0.017 & 0.844 & 0.427 & 0.658\\
            &DRD  & 19.4 & \bf{19.3} & \bf{0.229} & \bf{0.112} & \bf{0.066} & \bf{0.044} & 0.864 & 0.455 & 0.679\\
            & DialoGPT.   & 84.0 & 8.4 & 0.073 & 0.020 & 0.006 & 0.003 & 0.797 & 0.363 & 0.603\\
            &DialoGPT$_{finetune}$. & 16.2 & 19.0 & 0.165 & 0.076 & 0.040 & 0.023 & \bf{0.871} & \bf{0.461} & \bf{0.683}\\
            &GPT-2$_{finetune}$.   & \bf{15.0} & 14.4 & 0.140 & 0.049 & 0.020 & 0.010 & 0.860 & 0.417 & 0.649\\

            \midrule
            \midrule
            
            \multirow{5}{*}{Wizard-of-Wikipedia(Test Unseen)} 
            &TMN & 103.6 & 14.3 & 0.168 & 0.057 & 0.022 & 0.009 & 0.839 & 0.408 & 0.645\\
            &DRD  & 23.0 & \bf{17.9} & \bf{0.221} & \bf{0.102} & \bf{0.057} & \bf{0.037} & 0.862 & 0.444 & 0.671\\
            & DialoGPT.   & 85.9 & 8.1 & 0.071 & 0.019 & 0.006 & 0.002 & 0.792 & 0.362 & 0.596\\
            &DialoGPT$_{finetune}$. & 20.4 & 17.6 & 0.158 & 0.067 & 0.032 & 0.017 & \bf{0.869} & \bf{0.451} & \bf{0.674}\\
            &GPT-2$_{finetune}$.   & \bf{18.9} & 13.8 & 0.139 & 0.047 & 0.019 & 0.008 & 0.859 & 0.411 & 0.642\\
            
            \midrule
            \midrule
            \multirow{5}{*}{CMU-DoG} 
            &TMN & 75.2 & 9.9 & 0.115 & 0.040
 & 0.016 & 0.007 & 0.789 & 0.399 & 0.615\\
            &DRD  & 54.4 & 10.7 & 0.150 & 0.057 & 0.025 & 0.012 & 0.809 & 0.413 & 0.633\\
            & DialoGPT.   & 73.4 & 6.9 & 0.091 & 0.022 & 0.006 & 0.002 & 0.762 & 0.358 & 0.576\\
            &DialoGPT$_{finetune}$. & \bf{15.9} & \bf{13.7} & \bf{0.161} & \bf{0.064} & \bf{0.030} & \bf{0.015} & \bf{0.812} & \bf{0.430} & \bf{0.641}\\
            &GPT-2$_{finetune}$.   & 16.5 & 9.4 & 0.124 & 0.038 & 0.014 & 0.006 & 0.780 & 0.382 & 0.597\\
            \midrule
            \midrule
            
            \multirow{5}{*}{Topic-Chat-Freq} 
            &TMN & 30.3 & 16.5 & 0.176 & 0.079 & 0.041 & 0.025 & 0.891 & 0.444 & 0.693 \\
            &DRD  & 25.9 & 15.2 & \bf{0.203} & \bf{0.088} & \bf{0.050} & \bf{0.033} & 0.893 & 0.408 & 0.681\\
            & DialoGPT.   & 87.6 & 8.3 & 0.074 & 0.018 & 0.005 & 0.002 & 0.842 & 0.380 & 0.630\\
            &DialoGPT$_{finetune}$. & \bf{13.2} & \bf{17.1} & 0.161 & 0.069 & 0.036 & 0.020 & \bf{0.906} & \bf{0.446} & \bf{0.705}\\
            &GPT-2$_{finetune}$.   & 13.4 & 13.6 & 0.136 & 0.047 & 0.021 & 0.011 & 0.892 & 0.411 & 0.674\\
            \midrule
            \midrule
            \multirow{5}{*}{Topic-Chat-Rare} 
            &TMN & 52.1 & 14.6 & 0.168 & 0.068 & 0.031 & 0.016 & 0.881 & 0.429 & 0.682\\
            &DRD  & 28.0 & 15.1 & \bf{0.190} & \bf{0.083} & \bf{0.046} & \bf{0.030} & 0.874 & 0.398 & 0.667\\
            & DialoGPT.   & 87.9 & 8.5 & 0.071 & 0.018 & 0.005 & 0.002 & 0.835 & 0.372 & 0.623\\
            &DialoGPT$_{finetune}$. & \bf{15.7} & \bf{16.7} & 0.156 & 0.063 & 0.030 & 0.015 & \bf{0.900} & \bf{0.437} & \bf{0.697}\\
            &GPT-2$_{finetune}$.   & 16.2 & 13.1 & 0.130 & 0.042 & 0.017 & 0.008 & 0.884 & 0.401 & 0.664\\
            
            \bottomrule
        \end{tabular}
         }
        \caption{\label{tb:metric} Automatic evaluation results. Numbers in bold indicate the best performing models.}
    \label{tb2:auto_eval}
\end{table*}

\subsection{Analysis with Automatic Metrics}
We first examine the models with automatic metrics, including perplexity (PPL), BLEU \cite{papineni2002bleu}, $\{\text{Average}, \text{Extrema}, \text{Greedy}\}$ \cite{liu2016not}, and F1 \cite{dinan2018wizard}. The last metric is calculated with the code shared at  \url{https://github.com/facebookresearch/ParlAI/blob/master/parlai/core/metrics.py}, while the others are computed using a public NLG evaluation project available at \url{https://github.com/Maluuba/nlg-eval}.  We choose Transformer Memory Network (TMN) \cite{dinan2018wizard}
and Disentangled Response Decoder (DRD) \cite{zhao2020low} as baselines, where DRD exploits pre-training techniques to handle the low-resource challenge in knowledge-grounded dialogue generation, and holds the state-of-the-art performance on Wizard when all parameters are fine-tuned with the full training data. 


\begin{table*}[!t]
    \centering
    \small
	\scalebox{0.8}{
    \begin{tabular}{*{8}{c}}
        \toprule
        \multirow{2}{*}{Dataset} & \multirow{2}{*}{Models} & \multicolumn{2}{c}{Utterance} & \multicolumn{3}{c}{Knowledge} & \multirow{2}{*}{$\kappa$}\\
        \cmidrule(lr){3-4} \cmidrule(lr){5-7} && Fluency (\%) & Coherency (\%) & Existence (\%) & Relevance (\%) & Correctness (\%) \\
        \midrule
        \multirow{4}{*}{Test Seen}
        & DialoGPT. & 94.0 & 62.0 & 33.2 & 26.0 & 21.0 & 0.64  \\
        & DialoGPT$_{finetune}$. & 95.4 & 77.8 & 74.6 & 61.8 & 42.6 & 0.68 \\
        & GPT-2$_{finetune}$. & 96.8 & 80.4 & 77.4 & 67.0 & 47.6 & 0.61  \\
        & DRD  & 84.2 & 74.6 & 67.8 & 57.6 & 50.4 & 0.62 \\
        
        \midrule
        \midrule
        
        \multirow{4}{*}{Test Unseen}
        & DialoGPT. & 94.4 & 67.6 & 25.6 & 20.8 & 13.8 & 0.64  \\
        & DialoGPT$_{finetune}$. & 96.4 & 78.4 & 68.8 & 56.6 & 36.0 & 0.63 \\
        & GPT-2$_{finetune}$. & 97.0 & 82.2 & 74.6 & 64.2 & 41.4 & 0.61  \\
        & DRD  & 83.8 & 74.8 & 64.2 & 55.4 & 42.0 & 0.68 \\    
            
        \bottomrule
    \end{tabular}}
    \caption{Human evaluation results. The ratios are calculated by combining annotations from four judges together.}
    \label{tb3:human_eval}
\end{table*}

Table \ref{tb2:auto_eval} reports the evaluation results. Obviously,  DialoGPT is much worse than the baselines in terms of all metrics over all the three datasets, due to the gap between conversations in Reddit (at least those for training) and conversations in the probes. The results indicate that DialoGPT under a zero-shot setting is still not capable of going deep in discussions about a specific topic, even though its responses carry some content case-by-case as shown in \cite{zhang2019dialogpt}. However, this does not mean that there is ``no knowledge'' in the pre-trained language models, as after fine-tuning on the training sets of the probes, DialogGPT is comparable with
the baselines on most of the metrics. Note that we follow \cite{zhao2020low} and exploit a vocabulary with $60,000$ words for the baselines. Although the difference in vocabulary size is not significant, PPL is still not directly comparable between the pre-trained models and the baselines. In terms of F1, DialogGPT$_{finetune}$ is comparable with the baselines on Wizard, and outperforms the baselines on the other data, indicating that the model can leverage the knowledge in its parameters and reply with relevant and specific content along with topic of the dialogue contexts, at least judged from unigram overlap with human responses. The advantage of DialogGPT$_{finetune}$ over the baselines is even bigger on CMU$\_$DoG, indicating that pre-training is more crucial when training data is small in size.  GPT-2$_{finetune}$ is worse than DialogGPT$_{finetune}$, probably due to the small training size of the probes.

Overall, it seems that pre-trained language models can be used to ground open domain dialogues as long as we can find a few dialogues carrying knowledge for fine-tuning, though how to obtain such dialogues (e.g., without crowd-sourcing) for a specific task could arise as a new problem. On the other hand, it is well-known that there exist one-to-many relations in open domain dialogues at both a knowledge-level and a response-level. Therefore, the automatic evaluation may not be sufficient to support our conclusions, which motivates us to further examine the pre-trained language models with human annotations.

\subsection{Further Analysis with Human Judgment}
Since human labor is expensive, we only compare the pre-trained language models with DRD on Wizard. For each of Test Seen and Test Unseen, we randomly sample $500$ contexts. Responses from different models for these contexts are pooled, randomly shuffled, and presented to $4$ well-educated native speakers as annotators. The annotators judge the quality of each response from five aspects: fluency, coherence regarding the contexts, if the response carries knowledge (knowledge existence), if the knowledge is relevant with the contexts (knowledge relevance), and if the knowledge is correct (knowledge correctness). On each aspect, a binary label (0/1) is assigned to a response by an annotator, and in total each response receives $4$ labels per aspect. Annotators are encouraged to actively use search engines when they feel difficult to make a decision. Note that if an annotator judges there is no knowledge in a response, then by default he/she will also assign $0$s to the response on knowledge relevance and knowledge correctness. The agreement among the annotators is measured by Fleiss’ kappa \cite{fleiss1971measuring}.

Table \ref{tb3:human_eval} summarizes human evaluation results. All kappa values exceed $0.6$, indicating substantial agreement among the annotators. From the comparison, we obtain the following insights: (1)  responses from DialoGPT are generally grammatical, but they could digress from the contexts or be too generic sometimes. Moreover, the responses are more like casual chat rather than information exchange, as demonstrated by case-1 in Table \ref{case}; 
(2) after fine-tuning, both coherence and knowledge-related metrics get improved. The fine-tuned models are good at delivering commonsense knowledge, but often make mistakes on details such as numbers, time, locations, etc. Case-2 in Table \ref{case} is such an example. Another observation is that the fine-tuned models sometimes glue different pieces of knowledge together in an awkward way, causing disorder within responses, as demonstrated by case-3 in Table \ref{case}; and (3) though DRD sometimes generates responses with repetition/grammatical errors/irrelevant facts (due to wrong knowledge selection), it does a better job on details, thanks to the external knowledge. More cases are shown in the supplementary material. 

In conclusion, the human evaluation further convinces us that with proper fine-tuning material, pre-trained language models are knowledgeable to ground open domain dialogues. Despite this, combining external knowledge with the pre-trained language models could still be an important future direction (considering (2) and (3)).

 	   \section{Conclusions}
We inspect how knowledgeable pre-trained language models are in the task of open domain dialogue generation. Evaluation results on benchmarks indicate that by fine-tuning on a few dialogues carrying knowledge, a pre-trained language model can generate proper and informative responses without external knowledge sources.

\newpage
\clearpage
\bibliography{emnlp2020}
\bibliographystyle{acl_natbib}
\clearpage
\appendix

\section{More Details of the Benchmarks}
Table \ref{tbl:stat} reports some statistics of Wizard, CMU$\_$DOG, and Topic$\_$Chat.

\begin{table*}[ht]
\centering
\resizebox{\linewidth}{!}{
\begin{tabular}{c|c|c|c|c|c|c|c|c|c|c|c}
\hline
\multirow{2}{*}{}          & \multicolumn{4}{c|}{Wizard of Wikipedia}   & \multicolumn{3}{c|}{CMU$\_$DoG}  & \multicolumn{4}{c}{Topic$\_$Chat} \\ \cline{2-12}
                          & Train   & Valid  & Test Seen & Test Unseen & Train    & Valid  & Test & Train   & Valid  & Test Freq & Test Rare   \\ \hline
Number of Dialogues    & 18,430  & 1,948  & 965       & 968         & 3,373    & 229    & 619   & 8,628  & 1,078  & 539  & 539   \\  \hline
Average Utterances per Dialogue & 9.0     & 9.1    & 9.0       & 9.1         & 22.2     & 21.8   & 22.0  & 21.8  & 21.7  & 21.8  & 21.8  \\ \hline
Average Words per Utterance & 16.4     & 16.4    & 16.4       & 16.1         & 10.9   & 12.2   & 10.9  & 19.5  & 19.8  & 19.5  & 19.5    \\ 
\bottomrule
\end{tabular}
}
\caption{Statistics of the three datasets.}
\label{tbl:stat}
\end{table*}

\section{More Baselines}
Besides Transformer Memory Network (TMN) \cite{dinan2018wizard} and Disentangled Response Decoder (DRD) \cite{zhao2020low}, we also compare the pre-trained models with Incremental Transformer with Deliberation Decoder (ITDD) \cite{li2019incremental} and Posterior Knowledge Selection (PostKS) \cite{lian2019learning} on automatic metrics. Table \ref{tb2:stat} reports the evaluation results. Basically, the conclusions we have drawn in the main paper still hold here.

\begin{table*}[!ht]
    \small
    \centering
      \scriptsize
       \scalebox{0.9}{
        \begin{tabular}{clcccccccccccc}
            \toprule
            Dataset  & Model  & PPL & F1 & BLEU-1 & BLEU-2 & BLEU-3 & BLEU-4 & Average & Extrema  & Greedy  \\
            \midrule
            \multirow{5}{*}{Wizard-of-Wikipedia(Test Seen)}
            &ITDD  & 17.8 & 16.2 & 0.158 & 0.071 & \bf{0.041} & \bf{0.025} & 0.841 & 0.425 & 0.654\\
            &PostKS  & 68.5 & 17.9 & \bf{0.167} & \bf{0.076} & 0.040 & 0.024 & 0.774 & 0.406 & 0.625\\
            & DialoGPT.   & 84.0 & 8.4 & 0.073 & 0.020 & 0.006 & 0.003 & 0.797 & 0.363 & 0.603\\
            &DialoGPT$_{finetune}$. & 16.2 & \bf{19.0} & 0.165 & \bf{0.076} & 0.040 & 0.023 & \bf{0.871} & \bf{0.461} & \bf{0.683}\\
            &GPT-2$_{finetune}$.   & \bf{15.0} & 14.4 & 0.140 & 0.049 & 0.020 & 0.010 & 0.860 & 0.417 & 0.649\\

            \midrule
            \midrule
            
            \multirow{5}{*}{Wizard-of-Wikipedia(Test Unseen)} 
            &ITDD  & 44.8 & 11.4 & 0.134 & 0.047 & 0.021 & 0.011 & 0.826 & 0.364 & 0.624\\
            &PostKS  & 106.7 & 14.1 & 0.149 & 0.056 & 0.024 & 0.011 & 0.735 & 0.355 & 0.589\\
            & DialoGPT.   & 85.9 & 8.1 & 0.071 & 0.019 & 0.006 & 0.002 & 0.792 & 0.362 & 0.596\\
            &DialoGPT$_{finetune}$. & 20.4 & \bf{17.6} & \bf{0.158} & \bf{0.067} & \bf{0.032} & \bf{0.017} & \bf{0.869} & \bf{0.451} & \bf{0.674}\\
            &GPT-2$_{finetune}$.   & \bf{18.9} & 13.8 & 0.139 & 0.047 & 0.019 & 0.008 & 0.859 & 0.411 & 0.642\\
            
            \midrule
            \midrule
            \multirow{5}{*}{CMU-DoG} 
            &ITDD  & 26.0 & 10.4 & 0.095 & 0.036 & 0.017 & 0.009 & 0.748 & 0.390 & 0.587\\
            &PostKS  & 62.2 & 11.7 & 0.143 & 0.051 & 0.022 & 0.010 & \bf{0.818} & 0.424 & 0.637\\
            & DialoGPT.   & 73.4 & 6.9 & 0.091 & 0.022 & 0.006 & 0.002 & 0.762 & 0.358 & 0.576\\
            &DialoGPT$_{finetune}$. & \bf{15.9} & \bf{13.7} & \bf{0.161} & \bf{0.064} & \bf{0.030} & \bf{0.015} & 0.812 & \bf{0.430} & \bf{0.641}\\
            &GPT-2$_{finetune}$.   & 16.5 & 9.4 & 0.124 & 0.038 & 0.014 & 0.006 & 0.780 & 0.382 & 0.597\\
            \midrule
            \midrule
            
            \multirow{5}{*}{Topic-Chat-Freq} 
            &ITDD  & 21.4 & 15.8 & 0.163 & \bf{0.074} & \bf{0.041} & \bf{0.026} & 0.887 & 0.426 & 0.680\\
            &PostKS  & 48.0 & 15.4 & \bf{0.167} & 0.069 & 0.033 & 0.018 & 0.884 & 0.419 & 0.676\\
            & DialoGPT.   & 87.6 & 8.3 & 0.074 & 0.018 & 0.005 & 0.002 & 0.842 & 0.380 & 0.630\\
            &DialoGPT$_{finetune}$. & \bf{13.2} & \bf{17.1} & 0.161 & 0.069 & 0.036 & 0.020 & \bf{0.906} & \bf{0.446} & \bf{0.705}\\
            &GPT-2$_{finetune}$.   & 13.4 & 13.6 & 0.136 & 0.047 & 0.021 & 0.011 & 0.892 & 0.411 & 0.674\\
            \midrule
            \midrule
            \multirow{5}{*}{Topic-Chat-Rare} 
            &ITDD  & 24.7 & 14.0 & 0.153 & 0.062 & \bf{0.032} & \bf{0.019} & 0.880 & 0.408 & 0.670\\
            &PostKS  & 69.2 & 14.5 & 0.149 & 0.056 & 0.025 & 0.012 & 0.860 & 0.379 & 0.654\\
            & DialoGPT.   & 87.9 & 8.5 & 0.071 & 0.018 & 0.005 & 0.002 & 0.835 & 0.372 & 0.623\\
            &DialoGPT$_{finetune}$. & \bf{15.7} & \bf{16.7} & \bf{0.156} & \bf{0.063} & 0.030 & 0.015 & \bf{0.900} & \bf{0.437} & \bf{0.697}\\
            &GPT-2$_{finetune}$.   & 16.2 & 13.1 & 0.130 & 0.042 & 0.017 & 0.008 & 0.884 & 0.401 & 0.664\\
            
            \bottomrule
        \end{tabular}
         }
        \caption{\label{tb:metric} Automatic evaluation results}
    \label{tb2:stat}
\end{table*}

\section{More Cases}
Table \ref{case_1} shows more examples from DialoGPT where the responses are good in a chat style but lack necessary information for extending the discussions. In spite of this, we also see a few good cases from DialoGPT, as shown in Table \ref{case_2}, which remind us of the potential of pre-trained language models in terms of replying with knowledge.  Table \ref{case_3} and Table \ref{case_5} give more cases where DialogGPT$_{finetune}$ and GPT-2$_{finetune}$ do an excellent job respectively, and Table \ref{case_4} and Table \ref{case_6} show more examples where the two models make mistakes on details.

\begin{table*}[!ht]
		\small
		\centering	
		{		
			\begin{tabular}{p{2.5cm}|l}
				\hline
				\multirow{1}{*}{}   &\textbf{Case-1}\\ \hline
				\multirow{4}{*}{Context}   &\textbf{Utter1}: I suffer from Hypochondriasis, a verious serious disease 
				\\ &\textbf{Utter2}: I haven't heard of that, what's it like?\\ &\textbf{Utter3}: Is a condition in which a person is inordinately worried about having a serious illness. \\ &\textbf{Utter4}: Ah, ok, I have heard of that. Just didn't know the term. I'm sure that is frustrating. \\\hline
				\multirow{1}{*}{DialoGPT} &It's exactly what it is .\\ \hline
				\midrule
				\multirow{1}{*}{}   &\textbf{Case-2}\\ \hline
				\multirow{5}{*}{Context}   &\textbf{Utter1}: I love Iguanas, I have a few as pets. Do you like lizards at all? 
				\\ &\textbf{Utter2}: Yes, I like them. They are interesting.and prehistoric looking. I like turtles too.\\ &\textbf{Utter3}: I agree, they definitely have a prehistoric look to them. There are also over 6000 species \\ &\qquad \ \ \ \ \, spread across the world. \\ &\textbf{Utter4}: Do they have teeth and does their bite hurt if they bite you? \\\hline
				\multirow{1}{*}{DialoGPT} &yes it hurts .\\
                \hline
				\midrule
				\multirow{1}{*}{}   &\textbf{Case-3}\\ \hline
				\multirow{2}{*}{Context}   &\textbf{Utter1}: Hey have you ever been to Baton Rouge?
				\\ &\textbf{Utter2}: I haven't. Is it a nice play to live or visit?
				\\ \hline
				\multirow{1}{*}{DialoGPT} & It's a nice town .\\ \hline
				
			\end{tabular}
		}
		\caption{\label{case_1} Responses from DialoGPT are more like chat.}
	\end{table*}

\begin{table*}[ht]
		\small
		\centering	
		{		
			\begin{tabular}{p{2.5cm}|l}
				\hline
				\multirow{1}{*}{}   &\textbf{Case-1}\\ \hline
				\multirow{5}{*}{Context}   &\textbf{Utter1}: Sometimes I am overcome by laziness, and I don't want to work or do anything at all. 
				\\ &\textbf{Utter2}: I know exactly how you feel. I can get pretty lazy.\\ &\textbf{Utter3}: More formally some call laziness a disinclination to activity or exertion despite having the \\ &\qquad \ \ \ \ \,  ability to  \\ &\textbf{Utter4}: I am trying not to become too lazy. \\\hline
				\multirow{1}{*}{DialoGPT} &It's better to be lazy . It means you're more careful with your energy budget .\\ \hline
				\midrule
				\multirow{1}{*}{}   &\textbf{Case-2}\\ \hline
				\multirow{4}{*}{Context}   &\textbf{Utter1}: Instagram is a great way to share picture with my family and friends. Do you use it? 
				\\ &\textbf{Utter2}: I do not use Instagram, but I do use Facebook. Apparently Facebook owns Instagram.\\ &\textbf{Utter3}: Ya. That is why you get messages in FB about Instagram. I heard the younger people like \\ &\qquad \ \ \ \ \, Instagram better than FB. \\\hline
				\multirow{1}{*}{DialoGPT} &Younger crowd were in Facebook before Instagram was popular .\\
                \hline
				\midrule
				\multirow{1}{*}{}   &\textbf{Case-3}\\ \hline
				\multirow{3}{*}{Context}   &\textbf{Utter1}: I like to study, collect and race cars in my free time.
				\\ &\textbf{Utter2}: Same here ! Auto racing is also known as automobile racing \\ &\textbf{Utter3}:I think that sounds very old fashioned. lol. What cars do you race?
				\\ \hline
				\multirow{1}{*}{DialoGPT} & I'll drive a Maserati that I own and race my alfa one\\ \hline
				
			\end{tabular}
		}
		\caption{\label{case_2} Informative responses from DialoGPT.}
	\end{table*}	
	
\begin{table*}[!ht]
		\small
		\centering	
		{		
			\begin{tabular}{p{2.5cm}|l}
				\hline
				\multirow{1}{*}{}   &\textbf{Case-1}\\ \hline
				\multirow{4}{*}{Context}   &\textbf{Utter1}: I suffer from Hypochondriasis, a verious serious disease 
				\\ &\textbf{Utter2}: I haven't heard of that, what's it like?\\ &\textbf{Utter3}: Is a condition in which a person is inordinately worried about having a serious illness. \\ &\textbf{Utter4}: Ah, ok, I have heard of that. Just didn't know the term. I'm sure that is frustrating. \\\hline
				\multirow{1}{*}{DialoGPT$_{finetune}$.} &It can cause memory loss, anxiety and depression\\ \hline
				\midrule
				\multirow{1}{*}{}   &\textbf{Case-2}\\ \hline
				\multirow{4}{*}{Context}   &\textbf{Utter1}: I love Jane Austen don‘t you! 
				\\ &\textbf{Utter2}: She sure is, I have read all three volumes of her juvenile writings in manuscript and \\ &\qquad \ \ \ \ \, even her unfinished novel, The Watsons.\\ &\textbf{Utter3}: Do you know anything about her that might be of interest to me? \\\hline
				\multirow{2}{*}{DialoGPT$_{finetune}$.} &Yes she was a romance novel writer, that's why she wrote so many with romantic themes in \\ & them.
\\
                \hline
				\midrule
				\multirow{1}{*}{}   &\textbf{Case-3}\\ \hline
				\multirow{2}{*}{Context}   &\textbf{Utter1}: A taco can be made with a variety of fillings! I love including pork and cheese
				\\ &\textbf{Utter2}: What culture are tacos from?
				\\ \hline
				\multirow{2}{*}{DialoGPT$_{finetune}$.} & I think most people have a taco from Mexico, although some may be made with beef or pork or \\ & even fish!\\ \hline
				
			\end{tabular}
		}
		\caption{\label{case_3} Good cases of DialoGPT$_{finetune}$.}
	\end{table*}

\begin{table*}[!ht]
		\small
		\centering	
		{		
			\begin{tabular}{p{2.5cm}|l}
				\hline
				\multirow{1}{*}{}   &\textbf{Case-1}\\ \hline
				\multirow{2}{*}{Context}   &\textbf{Utter1}: Hey have you ever been to Baton Rouge? 
				\\ &\textbf{Utter2}: I haven't. Is it a nice play to live or visit?\\\hline
				\multirow{1}{*}{DialoGPT$_{finetune}$.} &It's the 32nd largest metropolitan area. You should visit! \textcolor{red}{(wrong number)}\\ \hline
				\midrule
				\multirow{1}{*}{}   &\textbf{Case-2}\\ \hline
				\multirow{4}{*}{Context}   &\textbf{Utter1}: Who even is Kurt Cobain? 
				\\ &\textbf{Utter2}: Kurt Cobain was an american singer, and musician lead singer of the band nirvana and \\ &\qquad \ \ \ \ \, quite a story\\ &\textbf{Utter3}: Ohh wow I didn't know that he was on Nirvana. \\\hline
				\multirow{1}{*}{DialoGPT$_{finetune}$.} &Yes he was, he was born and raised in Los Angeles, California \textcolor{red}{(wrong location)}\\
                \hline
				\midrule
				\multirow{1}{*}{}   &\textbf{Case-3}\\ \hline
				\multirow{3}{*}{Context}   &\textbf{Utter1}: I like to build tanks! An aquarium (plural: "aquariums" or "aquaria") is a vivarium of \\ &\qquad \ \ \ \ \, any size having at least one transparent side
				\\ &\textbf{Utter2}: Well, that is a pretty cool job! I bet it take a large amount of work to do? 
				\\ \hline
				\multirow{2}{*}{DialoGPT$_{finetune}$.} & It can, and would be just an aquarium. A word for an aquarium is "tidal boulevard." there are no \\ & identifying features such as plants, invertebrates, or fungi \textcolor{red}{(mixed knowledge)}\\ \hline
				
			\end{tabular}
		}
		\caption{\label{case_4} Bad cases of DialoGPT$_{finetune}$.}
	\end{table*}
	
\begin{table*}[!ht]
		\small
		\centering	
		{		
			\begin{tabular}{p{2.5cm}|l}
				\hline
				\multirow{1}{*}{}   &\textbf{Case-1}\\ \hline
				\multirow{4}{*}{Context}   &\textbf{Utter1}: I love chihuahuas, I have 2 of them. 
				\\ &\textbf{Utter2}: Wow chihuahua are something else. Tiny and the smallest breed of dog. \\ &\textbf{Utter3}: Yep, they get a bad rap about their demeanor too. Most people say they're mean but one of \\ &\qquad \ \ \ \ \, mine is the sweetest ever.\\\hline
				\multirow{1}{*}{GPT$_{finetune}$.} &Lol cool! Chihuahuas come from indigenous Mexican dogs.\\ \hline
				\midrule
				\multirow{1}{*}{}   &\textbf{Case-2}\\ \hline
				\multirow{5}{*}{Context}   &\textbf{Utter1}: The greenery in Ireland is quite lovely. 
				\\ &\textbf{Utter2}: It is beautiful. It's the second largest island in the British Isles. Great Britain is the largest.\\ &\textbf{Utter3}: That is good to know. What about the population? \\ &\textbf{Utter4}: It was about 6.6 million in 2011. It was ranked the second most populous island in Europe. \\ &\textbf{Utter5}: Will the Brexit movement change much there? \\\hline
				\multirow{1}{*}{GPT$_{finetune}$.} &I think so. Just before the 2008 financial crisis, Ireland was one of the fastest growing economies.\\
                \hline
				\midrule
				\multirow{1}{*}{}   &\textbf{Case-3}\\ \hline
				\multirow{2}{*}{Context}   &\textbf{Utter1}: I have never visited Chicago but I hear the University is a private research university \\ &\qquad \ \ \ \ \,  located  there.
				\\ &\textbf{Utter2}: Which university? \\ 
				\hline
				\multirow{1}{*}{GPT$_{finetune}$.} & Well the University of Chicago is one of the top and oldest universities in the United States.\\ \hline
				
			\end{tabular}
		}
		\caption{\label{case_5} Good cases of GPT$_{finetune}$.}
	\end{table*}
	
\begin{table*}[!ht]
		\small
		\centering	
		{		
			\begin{tabular}{p{2.5cm}|l}
				\hline
				\multirow{1}{*}{}   &\textbf{Case-1}\\ \hline
				\multirow{2}{*}{Context}   &\textbf{Utter1}: Do you drink coffee? I occasionally drink coffee because the caffeine is a great stimulant 
				\\ &\textbf{Utter2}: I love coffee! Living in the south, I'm partial to iced coffee. I particularly like cold brew. \\\hline
				\multirow{1}{*}{GPT$_{finetune}$.} &My favourite is arabica. It's made from the seeds of the erythrocopus macrocarpa \textcolor{red}{(wrong fact)}\\ \hline
				\midrule
				\multirow{1}{*}{}   &\textbf{Case-2}\\ \hline
				\multirow{2}{*}{Context}   &\textbf{Utter1}: Do you like Elvis Presley? he was a famous american singer.
				\\ &\textbf{Utter2}: He is pretty good, I am from Nashville, So i've even been to graceland. \\\hline
				\multirow{1}{*}{GPT$_{finetune}$.} &He moved to Nashville, Tennessee when he was 14 in 1926. \textcolor{red}{(wrong time)}\\
                \hline
				\midrule
				\multirow{1}{*}{}   &\textbf{Case-3}\\ \hline
				\multirow{2}{*}{Context}   &\textbf{Utter1}: Man, ICP or Insane Clown Posse are my favorite band! Their hip hop style is phenomenal. \\ &\qquad \ \ \ \ \,  Do you like them?
				\\ &\textbf{Utter2}: I find ICP to have some of the worst fans, they seem to be crazy druggies quite often \\ 
				\hline
				\multirow{2}{*}{GPT$_{finetune}$.} & That's interesting because they're a punk rock band. I loved their second album "Stand Up" a lot. \\ & \textcolor{red}{(wrong fact)}\\ \hline
				
			\end{tabular}
		}
		\caption{\label{case_6} Bad cases of GPT$_{finetune}$.}
	\end{table*}

\end{document}